\documentclass[11pt,a4paper]{article}
\usepackage[hyperref]{naaclhlt2018}
\usepackage{times}
\usepackage{latexsym}
\usepackage{graphicx}
\usepackage{subfigure}

\usepackage{url}

\aclfinalcopy 

\title{Human vs Automatic Metrics: on the Importance of Correlation Design}

\author{Anastasia Shimorina \\
  Universit{\'e} de Lorraine / LORIA, Nancy, France \\
  {\tt anastasia.shimorina@loria.fr}
  }

\date{}

\begin{document}
\maketitle
\begin{abstract}
This paper discusses two existing approaches to the correlation analysis between automatic evaluation metrics and human scores in the area of natural language generation. Our experiments show that depending on the usage of a system- or sentence-level correlation analysis, correlation results between automatic scores and human judgments are inconsistent.
\end{abstract}

\section{Context and Motivation}
This work seeks to gain more insight into existing approaches to correlation analysis between automatic and human metrics in the area of natural language generation (NLG).

In the machine translation community, the practice to compare system- and sentence-level correlation results is well established \cite{callisonburchWMT08, callisonburchWMT09}. System-level analysis is motivated by the fact that automatic evaluation metrics such as \textsc{bleu} \cite{papineni2002bleu}, \textsc{meteor} \cite{denkowski2014meteor}, \textsc{ter} \cite{snover2006ter} were initially created to account for the evaluation of whole systems (i.e. they are corpus-based metrics). On the other hand, correlation analysis at the sentence level is motivated by the need to gauge the quality of individual sentences and more generally by the need to have a more fine-grained analysis of the results produced \cite{kulesza2004learning}. The common finding in MT is that automatic metrics correlate well with human judgments at the system level but much less so at the sentence level. This in turn prompted the search for alternative automatic metrics which would correlate well with human judgments at the sentence level. 

In NLG, there is a lack of such comparative studies. Traditional NLG evaluations and challenges (\citet{reiter2009investigation, gatt10introducing}, among others) used only system-level comparisons and reported low to strong correlations depending on the automatic metric used. \citet{reiter2009investigation} explicitly wrote that they did not compute correlations on individual texts because \textsc{bleu}-type metrics ``are not intended to be meaningful for individual sentences". 

Nonetheless, when researchers have one or few systems to evaluate, they resort to sentence-level correlation analysis: e.g., reports of \citet{stent2005evaluating} for paraphrasing, \citet{elliott2014comparing} for image caption generation. They usually report low to moderate correlations. Lately, \citet{novikova2017} observed the difference between results depending on the evaluation design but only report sentence-level correlation results as they have few systems in their study.

In the recent survey on the state of the art in NLG, \citet{gattkrahmer18survey} made a comparison of various validation studies, concluding that the studies yielded inconsistent results. However, it was not mentioned that the underlying design of those studies was different (some of them were system-based, other were sentence-based).

By this paper, we hope to raise awareness of different design in correlation analysis for NLG evaluation.
In this study, we present both a system- and a sentence-level correlation analysis on NLG data. We show that the results are similar to those obtained for MT systems and we conclude with some recommendations concerning the evaluation of NLG systems. 

\begin{figure*}[thp]
  \centering
  \subfigure[Spearman's $\rho$ at the system level. Crossed squares indicate that statistical significance was not reached ($\alpha = .05$). Human vs automatic metrics are in the black square.]{\includegraphics[scale=0.32]{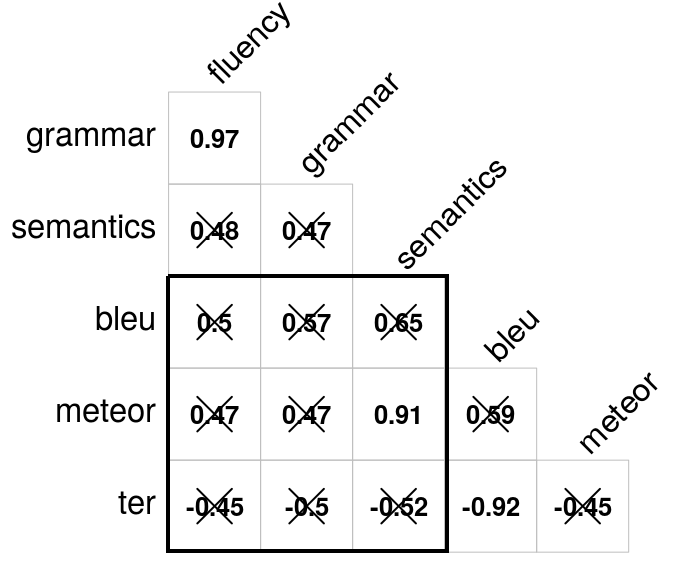}\label{fig:spearman-system}}
  \quad
  \subfigure[Spearman's $\rho$ at the sentence level. All correlations are statistically significant ($\alpha = .001$). Human vs automatic metrics are in the black square.]{\includegraphics[scale=0.32]{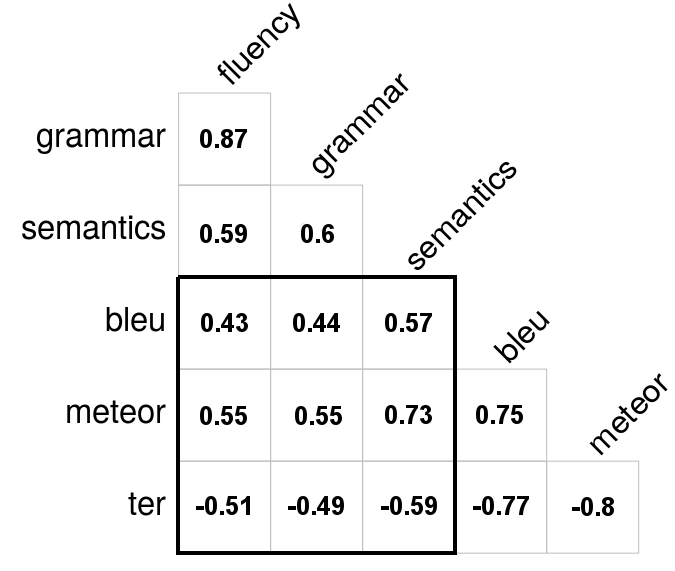}\label{fig:spearman-sentence}}
  \vspace{-0.3cm}
 \caption{System- and sentence-level correlation analysis.}\label{fig:spearman}
\end{figure*}

\section{Experimental Setup}
We used the \textsc{webnlg} dataset (\citet{gardent2017creating}) for our experiments. The dataset maps data to text, where a data input is a set of triples extracted from DBpedia, and a text is a verbalisation of those triples. We sampled 223 data inputs from \textsc{webnlg}, and used the outputs of nine different NLG systems which participated in the WebNLG Challenge \cite{gardent2017challenge}\footnote{\url{https://webnlg-challenge.loria.fr/challenge_2017/\#challenge-results}}.

The data inputs were chosen based on different characteristics of the \textsc{webnlg} corpus: how many RDF triples were in data units (size from 1 to 5), and what was the DBpedia category (Building, City, Artist, etc.). A sample for each system comprised texts from each category (15 texts); in each category all triple set sizes were covered (5 sizes), and finally we extracted 3 texts per every category and every size\footnote{One should note that, in such a way, our sample should have had 225 (i.e. $15*5*3$) texts; however, the count was reduced to 223, as one category (ComicsCharacter) had few data units for a particular size.}.

Automatic evaluation results (i.e., \textsc{meteor}, \textsc{ter}, \textsc{bleu-4} scores) were calculated for each NLG system both at the system and at the sentence-level and comparing each generated sentence against three references on average.

We crowdsourced\footnote{We used CrowdFlower to run evaluation and \textsc{mace} \cite{hovy2013learning} to remove unreliable judges.} human judgments. Each participant was asked to rate each generated text on a a three-point Lickert scale for semantic adequacy, grammaticality and fluency. We collected three judgments per text.

To perform correlation analysis both at system and sentence levels, we used Spearman's correlation coefficient\footnote{All statistical experiments were conducted using R. Data and scripts are available at \url{https://gitlab.com/shimorina/webnlg-human-evaluation}}. To prevent a possible bias, we excluded human references from the analysis as their automatic scores are equal to 1.0 (\textsc{bleu}, \textsc{meteor}) and 0.0 (\textsc{ter}). Thus, for system-level analysis, we have nine data points to build a regression line.

\section{Correlation Analysis Results}

At the system level (Figure \ref{fig:spearman-system}), the only statistically significant correlation ($p < .001$) is between \textsc{meteor} and semantic adequacy\footnote{We are focusing here mostly on the correlation between human and automatic metrics as delineated by the black square in Figure \ref{fig:spearman}.}. Similar findings for \textsc{meteor} were reported in the MT community \citep{callisonburchWMT09} and in the image caption generation domain \citep{bernardi2016automatic}. We also found a strong correlation between \textsc{ter} and \textsc{bleu}, and between grammaticality and fluency judgments.

At the sentence-level, on the other hand (Figure \ref{fig:spearman-sentence}), all correlations are statistically significant ($p < .001$). The highest correlation between human and automatic metrics is between \textsc{meteor} and semantic adequacy ($\rho = 0.73$). For other human/automatic correlation results, the correlation is moderate, ranging from $\rho = 0.43$ to $\rho = 0.59$ in absolute numbers. Automatic metrics show strong correlations with each other ($\rho \geq 0.78$).

In sum, there is a strong discrepancy between system- and sentence-level correlation results. Significance was not reached for most of the system-level correlations. At the sentence level, all correlations are significant, however the correlation between automatic metrics and human scores remains relatively low thereby confirming the findings of the MT community. At the sentence level, statistical significance is easier to achieve, since there are more data points than for the system-level analysis. One possibility to have statistically significant results at the system-level would be to use one-tailed test (instead of two-tailed) without a Bonferroni multiple-hypothesis correction, as it was done by \citet{reiter2009investigation}. However, that test is considered less statistically robust.

\section{Conclusion}
We argued that in NLG, as in MT, the specific type (system- vs sentence-level) of correlation analysis chosen to compare human and automatic metrics strongly impacts the outcome. While system-level correlation analyses have repeatedly been used in NLG challenges, sentence-level correlation is more relevant as it better supports error analysis. Based on an experiment, we showed that, in NLG as in MT, the sentence-level correlation between human and automatic metrics is low which in turn suggests the need for new automatic evaluation metrics for NLG that would better correlate with human scores at the sentence level.  

\bibliography{winlp2018}
\bibliographystyle{acl_natbib}

\end{document}